\documentclass[runningheads]{llncs}

\usepackage{eccv}

\usepackage{eccvabbrv}

\usepackage{graphicx}
\usepackage{booktabs}
\usepackage{todonotes}
\usepackage{mathtools}
\usepackage{fancyvrb}

\usepackage[accsupp]{axessibility}  

\usepackage{hyperref}

\usepackage{orcidlink}

\begin{document}

\title{Pix2Pix-OnTheFly: Leveraging LLMs for Instruction-Guided Image Editing} 

\titlerunning{Pix2Pix-OnTheFly}

\author{Rodrigo Santos\and João Silva \and António Branco}

\authorrunning{Santos et al.}

\institute{University of Lisbon\\ NLX - Natural Language and Speech Group, Department of Informatics\\ Faculdade de Ciências, Campo Grande, 1749-016 Lisboa, Portugal\\
\email{\{rsdsantos,jrsilva,antonio.branco\}@fc.ul.pt}}

\maketitle

\begin{abstract}
The combination of language processing and image processing keeps attracting increased interest given recent impressive advances that leverage the combined strengths of both domains of research. Among these advances, the task of editing an image on the basis solely of a natural language instruction stands out as a most challenging endeavour.

While recent approaches for this task resort, in one way or other, to some form of preliminary preparation, training or fine-tuning, this paper explores a novel approach: We propose a preparation-free method that permits instruction-guided image editing on the fly. 

This approach is organized along three steps properly orchestrated that resort to image captioning and DDIM inversion, followed by obtaining the edit direction embedding, followed by image editing proper. While dispensing with preliminary preparation, our approach demonstrates to be effective and competitive, outperforming recent, state of the art models for this task when evaluated on the MAGICBRUSH dataset.
  
\end{abstract}

\begin{figure}[h!]
    \centering
    \includegraphics[width=0.89\linewidth]{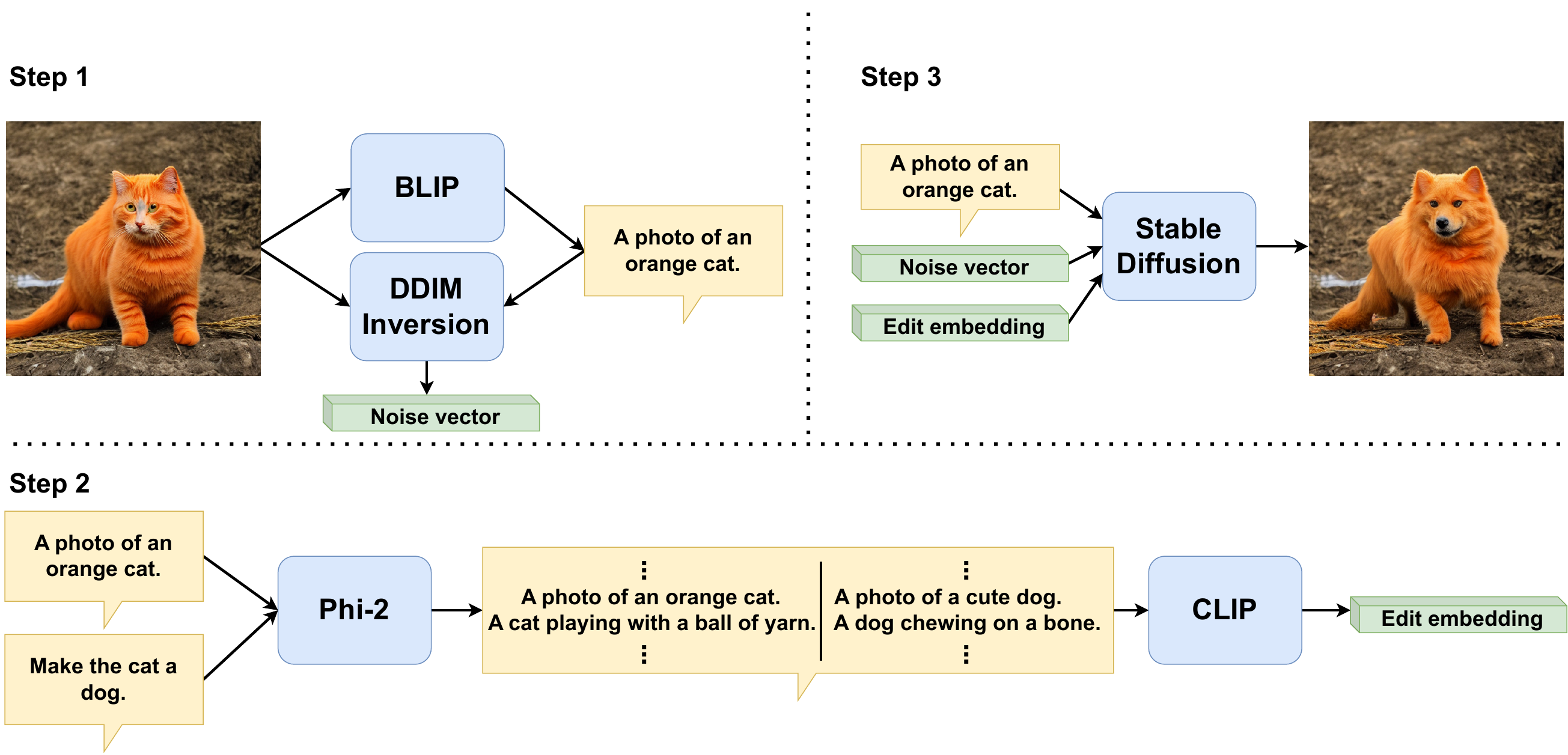}
    \caption{Overview of the architecture of our system}
    \label{fig:architecture}
\end{figure}

\begin{figure}[t]
    \centering
    \includegraphics[width=1\linewidth]{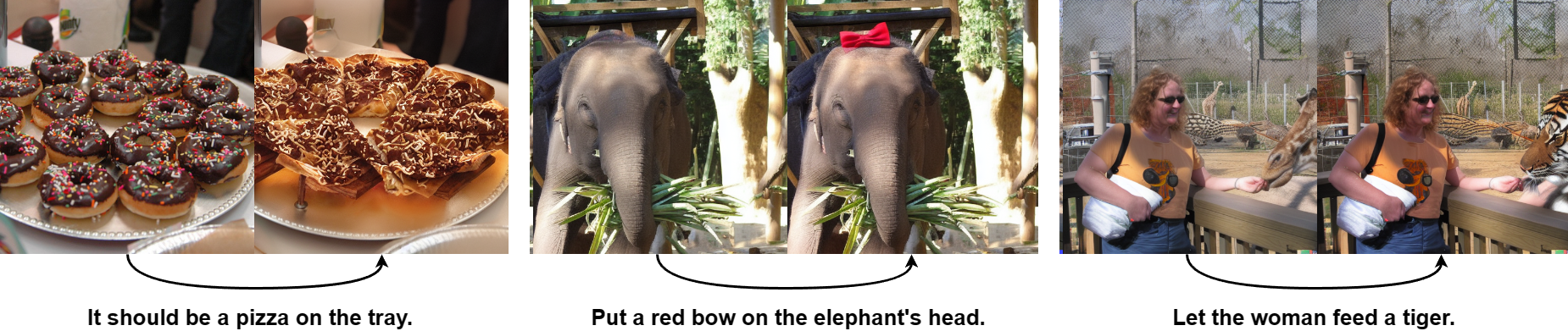}
    \caption{Instruction-guided image editing with our Pix2Pix-OnTheFly approach.}
    \label{fig:p2p_otf_examples}
\end{figure}

\section{Introduction}
\label{sec:intro}



In Artificial Intelligence, with its fast pace towards enhanced multimodal capabilities, the merger of language processing and image processing keeps attracting renewed interest and research given the fast progress increasingly leveraged by the reciprocal synergies between these two sub-fields.
In this context, the task of editing images solely through natural language requests or instructions emerges as a most enticing challenge, by means of which the expressive power of language is put at the service of a virtually unbounded handling and transformation of images.
While some approaches have focused on more constrained approaches of editing images through the alteration of image descriptions \cite{hertz:2022:prompt2prompt}, or through keyword based alterations \cite{Zhuang:2021:EditingLatentSpace}, we will address the more open and challenging task of editing images through requests freely expressed by means of unconstrained natural language expressions (see \cref{fig:p2p_otf_examples} for a few examples with our approach). 

Central to our approach is the adoption of cutting-edge neural architectures from these two sub-fields, namely a Diffusion model, exemplified by Stable Diffusion \cite{rombach:2022:stable_diffusion}, and a Language model, exemplified by the 2.7 billion-parameter Phi-2 \cite{gunasekar:2023:phi}.
And unlike conventional methods that rely on explicit training regimes, distinctive to our approach is the adoption of a training-free paradigm.
By seamlessly integrating the meaning representation of textual instructions into the image editing pipeline, we aim to unleash the whole process, allowing users to articulate their desired modifications with unconstrained linguistic elaboration, and without previous pre-processing stages of specific preparation or training.

We evaluate our proposed approach by using it to tackle the MAGICBRUSH dataset~\cite{zhang:2024:magicbrush}.
It outperforms previous state-of-the-art models specifically trained for this task, including from the InstructPix2Pix\cite{brooks:2023:instructpix2pix} and HIVE\cite{zhang:2023:hive} families.\footnote{Code is available at <\emph{removed for anonymous submission>}}

The remainder of this document is structured as follows.
\cref{sec:rel_work} covers related work and \cref{sec:approach} describes the novel approach used in this study.
\cref{sec:data} presents the data set used for testing, while \cref{sec:experiments} explains the experiments performed.
The results obtained are presented in \cref{sec:evaluation} and discussed in \cref{sec:discussion}.
Finally, \cref{sec:conclusion} closes the paper with concluding remarks.


\section{Related Work}
\label{sec:rel_work}

Historically, the fields of image and language processing have evolved independently, each dedicated to analyzing and generating its respective modality and to exploring the diverse applications stemming from their respective capabilities. 

In language processing, neural networks began achieving notable success through Recurrent Neural Networks applied to Machine Translation~\cite{sutskever2014sequence}.
However, new architectures and techniques, particularly Attention~\cite{bahdanau:2014:attention}, eventually showed improved performance, culminating in the Transformer~\cite{Vaswani:2017:Transformer}, which became the present mainstream architecture giving its clear superior performance in virtually any language processing task.

In image processing, in turn, the application of deep learning techniques to image generation has witnessed significant progress, namely with the emergence of the Generative Adversarial Network (GAN)~\cite{Goodfellow:2014:GANs}, consisting of a generator network tasked with creating realistic images and a discriminator network trained to distinguish between real and generated images~\cite{xu2017attngan,Zhu2019DMGANDM,tao2021dfgan}, thus iteratively enhancing their performance through adversarial competition.

While initially leveraging Convolutional Neural Networks~\cite{Lecun:1989:CNN}, recent advancements in GANs have explored alternative architectures, including Transformers \cite{Jiang:2021:TransGAN}, given their outstanding success in natural language processing. These advancements have unveiled promising prospects for advancing cross-modal processing, inducing a significant departure from the previous historical separation of these two sub-fields.



In the domain of image-to-language processing, notable advances have been made in tasks such as image captioning, involving the generation of textual descriptions for input images, and the supplementary task of image retrieval based on textual descriptions \cite{Radford:2021:CLIP, Xu:2015:Captioning,Wu:2017:Captioning1,Hossain:2019:Captioning2,Reed:2016:Generative,Guo:2018:shoesdataset,Yu:2017:relative1,Kovashka:2012:relative2}.

Conversely, in the language-to-image direction, progress has predominantly occurred in Conditional Image Generation tasks.
Two notable examples are DALL-E~\cite{Ramesh:2021:DALLE} and NÜWA~\cite{wu:2021:NUWA}, both resorting to Transformers to generate images based on textual prompts, albeit with distinct attention mechanisms and capabilities.
Additionally, CLIPGLASS \cite{Galatolo:2021:CLIPGLASS} demonstrates promising results in image generation by leveraging pre-trained Transformer models, showcasing the versatility of Transformer-based architectures.
Recent improvements have materialized in models like DALL-E~2 \cite{Ramesh:2022:DALLE2} and Stable Diffusion \cite{Dhariwal:2021:diffusion}, which incorporate the CLIP\cite{Radford:2021:CLIP} model for image and caption representation, along with diffusion models for image generation.

The present proposal differs from the ones mentioned above as it deals rather with the image-and-language to image direction: this task receives an image and a request as input in order to generate a new image, resulting from the input image having been changed in accordance to the request.

In this realm of language-driven image editing, the work by Cheng \etal~\cite{Cheng:2020:IIE} and Jiang \etal~\cite{Jiang:2021:LangGlobalEdit} explore the task by using a Generator/Discriminator architecture, and achieve good results despite the narrow domain of their datasets.

More recently, Hertz \etal~\cite{hertz:2022:prompt2prompt} developed an approach capable of editing images through the injection of cross-attention maps during the diffusion process.
With this method, the authors of InstructPix2Pix~\cite{brooks:2023:instructpix2pix} automatically create an image-editing dataset which they use to train a Stable Diffusion model for the task at hand.
Similarly, the authors of HIVE \cite{zhang:2023:hive} also train a Stable Diffusion model on a dataset of edited images with human feedback created by them.

While not guided by natural language requests,  the work of Parmar \etal~\cite{parmar:2023:pix2pix-zero} also extends the capabilities of a diffusion model for image editing.
These authors make use of pre-computed captions in order to create a fixed set of image alteration embeddings that can be applied to an image during the diffusion process culminating on an edited image.
This proposal is integrated in our approach, as it will be better detailed in \cref{sec:approach}.

\section{Approach}
\label{sec:approach}

We propose a novel neural framework for the task of image editing through natural language requests.
Building upon the contribution of Parmar \etal~\cite{parmar:2023:pix2pix-zero}, it seamlessly integrates the capability of instruction-guided editing into a model that previously only accepted pre-computed edit directions.

Our approach is divided into three steps (see \cref{fig:architecture}) that leverage different pre-trained models, namely
(i)~a diffusion model, Stable Diffusion~\cite{rombach:2022:stable_diffusion};
(ii)~a captioning model, BLIP \cite{li:2022:blip};
and (iii)~a large language model, Phi-2~\cite{gunasekar:2023:phi}.
This approach allows users to modify images based on textual instructions without the need for training or fine-tuning, solely relying on the emergent capabilities of these models when appropriately orchestrated together.
These steps are described below.

\subsection{Step 1: Image Captioning and DDIM Inversion}
\label{sec:approach_step1}
Caption conditioned diffusion models, such as Stable Diffusion, require both a caption describing the image to be generated and a starting noise vector.

To edit an image using diffusion, in turn, those two inputs are also required.
However, in an image editing scenario, apart from the edit request, only the initial image is provided.
Accordingly, it is necessary to first obtain both the caption and the noise vector that would produce that image when fed to the overall diffusion process.

\subsubsection{Captioning}
By the very nature of the task, the initial image is provided by the user, but its corresponding caption, or textual prompt in a text-to-image generation setup, is not.
To obtain one of the many possible prompts that could lead to the generation of the given image, we resort to a captioning model, BLIP \cite{li:2022:blip}.
Based on a Vision Transformer \cite{Dosovitskiy:2020:VIT}, this model is capable of Visual Question Answering, Image-Text Retrieval, and Image Caption.
We make use of the latter capability to generate the caption of the input image.

\subsubsection{DDIM Inversion}
With the initial image and its generated caption as input, we can then obtain the corresponding noise vector.
The process that accomplishes this is named DDIM (Denoising Diffusion Implicit Models) Inversion \cite{song:2020:DDIM}.
It consists of performing DDIM sampling in reverse order, \ie~applying the diffusion process going from image to noise.
Note that DDIM Inversion is lossy, as illustrated by the contrast between the first and second images in \cref{fig:dataset_example}.
We use 100 DDIM steps to obtain the noise vectors, the same as \cite{parmar:2023:pix2pix-zero}.

\subsection{Step 2: Obtaining the edit direction embedding}
\label{sec:approach_step2}

The second step focuses on obtaining the edit direction embedding that guides the diffusion process in order to successfully edit an image.

An edit direction embedding can be obtained with the help of the embedding model used by the image generator diffusion model, which is CLIP in this case.
For this purpose, a minimum of two textual captions are needed, namely a caption $c_\text{before}$ representing the image before editing and a caption $c_\text{after}$ representing the image after the desired transformation has been applied.
As both the final embedding and the initial embedding are vectors, their difference gives another vector, which can be interpreted as a direction from the initial image, before the edit, to the final image, after the edit.
Accordingly, the edit direction embedding $e_{\text{edit}}$ can be obtained by:
\begin{gather}
e_{\text{before}} = \text{CLIP}(c_\text{before}) \qquad e_{\text{after}} = \text{CLIP}(c_\text{after})\\
e_{\text{edit}} = e_{\text{after}}-e_{\text{before}}
\label{eq:edit_direction}
\end{gather}

As a refinement to this approach, a set with multiple before-edit captions and a set with multiple after-edit captions can be used, an average embedding calculated for each set, and the difference calculated from these two averages.

In the work of Parmar \etal~\cite{parmar:2023:pix2pix-zero}, the authors create these embeddings by asking GPT-3~\cite{Brown:2020:GPT3} to generate thousands of before-edit and after-edit captions.\footnote{For example, if the alteration request asked to add sunglasses to a cat, GPT-3 would be asked to generate thousands of before-edit captions about cats and thousands of after-edit captions about cats with sunglasses.}
This labour intensive task---both for machine and user---limits the number of possible edits the model can perform, only allowing for a restricted set of pre-computed edit directions since every new edit direction would need to be compiled.

In this work, a key innovation consist in generating these edit directions on-the-fly and in allowing for editing requests input by the user, \ie~these edit direction are generated with the textual request for image editing as input.

In order to accomplish this, we make use of a Large Language Model (LLM), namely Phi-2 \cite{gunasekar:2023:phi}, to generate before-edit and after-edit captions given an edit request.
We choose this model due to its compact size (2.7 billion parameters) and high performance, rivaling and even outperforming several larger LLMs such as Falcon \cite{Almazrouei:2023:falcon}, LLaMA2 \cite{touvron:2023:llama2} and Mistral \cite{jiang:2023:mistral}.\footnote{According to the HuggingFace leaderboard, Phi-2 has an average across tested tasks of 61.33, Falcon has 44.17, LLaMA2-7b 50.97, LLaMA-13b 55.69, and Mistral 60.97.} 

\begin{figure}[p]
\begin{Verbatim}[frame=single,fontsize=\small]
Instruct: Given the transformation `[TRANSFORMATION]', generate [NUMBER]
image captions for before and after the transformation.
Output: Before transformation

Caption 1:
\end{Verbatim}
\vspace{-3ex}
    \caption{Template of the prompt given to the language model for caption generation, where [TRANSFORMATION] is to be replaced with the edit request and [NUMBER] with the number of before-edit and after-edit captions to generate.}
    \label{fig:prompt_template}
\end{figure}

\begin{figure}[p]
\begin{Verbatim}[frame=single,fontsize=\small]
Instruct: Given the transformation `Make the cat a dog', generate 2
image captions for before and after the transformation.
Output: Before transformation

Caption 1:
\end{Verbatim}
\vspace{-3ex}
    \caption{Example of a prompt given to the language model for caption generation after the template has been instantiated with concrete values for the edit request and the number of captions to generate.}
    \label{fig:prompt_template_filled}
\end{figure}

\begin{figure}[p]
\begin{Verbatim}[frame=single,fontsize=\small]
Instruct: Given the transformation `Make the cat a dog', generate 2
image captions for before and after the transformation.
Output: Before transformation

Caption 1: A photo of a tabby cat sleeping.
Caption 2: A cat playing with a ball of yarn.

After transformation

Caption 1: A photo of a cute dog.
Caption 2: A dog chewing on a bone.
\end{Verbatim}
\vspace{-3ex}
    \caption{Example of the output generated by the language model.}
    \label{fig:prompt_example}
\end{figure}

\begin{figure}[p]
\begin{Verbatim}[frame=single,fontsize=\small]
Instruct: Given the transformation `Make it a dog', generate 2
image captions for before and after the transformation.
Output: Before transformation

Caption 1: A photo of an orange cat.
\end{Verbatim}
\vspace{-3ex}
    \caption{Example of a prompt given to the language model for caption generation, with the first before-edit caption already locked in.}
    \label{fig:prompt_lockin}
\end{figure}

\cref{fig:prompt_template} shows the template of the input given to the language model.
This input contains a prompt explaining the task to be performed.
The prompts of the Phi-2 model use ``Instruct:'' to mark the beginning of the instruction and ``Output:'' to mark where the model should start generating.
Note that in order to guide the generation and constraint the format of the output, in the prompt we provide the beginning of the output, namely with ``Before transformation'' and ``Caption~1:''.

The prompt allows for the control of what before-edit and after-edit captions need to be generated by replacing ``[TRANSFORMATION]'' with the desired textual editing request, as well as the number of before-edit and after-edit captions to be generated by replacing ``[NUMBER]'' with the desired number.
\cref{fig:prompt_template_filled} shows an example of a prompt that could be obtained from the template and \cref{fig:prompt_example} shows an example of the final generated result.

After the specified number of before-edit and after-edit captions have been generated, the edit direction embedding is calculated as in \cite{parmar:2023:pix2pix-zero}, using the method explained in \cref{eq:edit_direction}.

\subsection{Step 3: Image editing}
\label{sec:approach_step3}

The third and final step consists of generating the edited image.
This is done by generating a new image using Stable Diffusion.
However, this new image is conditioned by the noise and caption obtained in Step~1 and by the edit direction embedding obtained in Step~2.

This is similar to the process in \cite{hertz:2022:prompt2prompt}, where editing is guided by the injection of cross-attention maps during the diffusion process.
In our approach, this guidance is provided by the edit direction embedding obtained in Step~2, and also by adding this embedding to the embedding of the caption obtained in Step~1.

We use 100 DDIM steps to generate the new image, the same as \cite{parmar:2023:pix2pix-zero}.

\section{Data}
\label{sec:data}

\begin{figure}[t]
    \centering
    \includegraphics[width=1\linewidth]{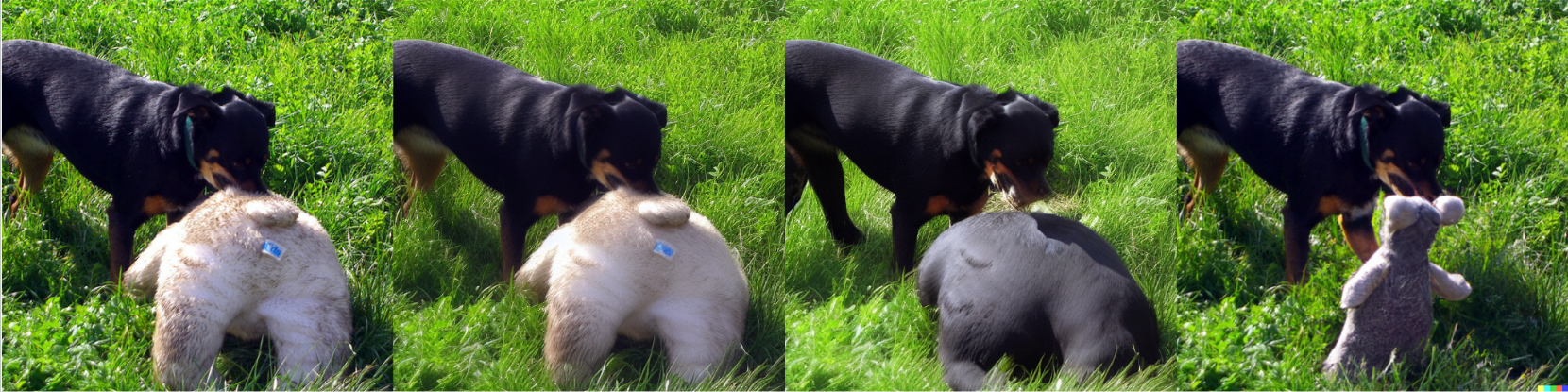}
    \caption{Example from the MAGICBRUSH test set.
    The edit request is ``Make the teddy bear black''.
    The four images are: the original image, the image generated from the noise obtained through DDIM Inversion, the image generated by our system, and the gold edited image from the dataset.}
    \label{fig:dataset_example}
\end{figure}

We use the MAGICBRUSH dataset~\cite{zhang:2024:magicbrush} to assess the performance of our model.
This dataset has been created through crowdsourcing, with workers from Amazon Mechanical Turk, with the help of the DALL-E~2 model, and it is based on images from the MSCOCO dataset \cite{lin:2014:coco}.

The dataset creation process involved workers proposing a textual edit instruction for each image, as well as the corresponding caption after applying the transformation to the image.
Workers were also asked to draw a free-form region mask on the image marking the area meant to be affected by the edit.
The masked image and the caption would be given to the DALL-E 2 model to generate a new image, through mask infilling, with the desired transformation. 

The final dataset consists of more than 10~thousand edit triples from which we use the 1053 triples that were left aside for testing purposes.

\subsubsection{Masking issues}

While, to the best of our knowledge, this dataset is the one with the highest quality for the task, it is not without its problems.
As seen in \cref{fig:dataset_example}, the teddy bear in the original image (first image) does not have the same shape as the teddy bear in the gold edited image (fourth image), despite the editing request only asking for a color change (``Make the teddy bear black'').

This problem arises from the masking that is performed on the image during dataset creation.
By masking the image, important information is lost, such as the specific shape of the initial object.
Since the editing was done through mask infilling, the DALL-E~2 model only needs to generate a new object that fits the description ``teddy bear'', thus possibly losing important features of the initial object not transmitted through this textual description.

\section{Experimental space}
\label{sec:experiments}

A key innovative aspect of our approach concerns the creation of the edit direction embeddings by means of the generation of before-edit and after-edit captions for a given textual edit request.
As such, we study the impact of various setups that affect the generation of these captions.

\subsection{Prompt detail}
A critical factor influencing this generation process is the prompt given to the model.
The task of prompt engineering has gained significant importance and attention, particularly with the rise of large language models capable of executing diverse tasks beyond their original training.
Studies such as \cite{yang:2023:llm_optimizers} concentrate on enhancing prompts to maximize performance across new tasks.

Taking this into account, we experiment with two different prompts with different degrees of detail in how they describe the task:
\begin{itemize}
    \item A terse prompt:
    ``Given the transformation `[TRANSFORMATION]' generate [NUMBER] image captions for before and after the transformation.'';
    \item and a more expressive and detailed prompt:
    ``Employing the specified method `[TRANSFORMATION]', craft [NUMBER] pairs of descriptive captions delineating the images both prior to and following the application of the transformation process, elucidating the changes brought about.''.
\end{itemize}

\subsection{Number of generated captions}

The approach of Parmar \etal~\cite{parmar:2023:pix2pix-zero} relies on the generation of ``a large bank of diverse sentences for both source $s$ and the target $t$'' (\ie~what we call the before-edit and after-edit captions).
The specific number of sentences generated is not given in the paper, but the corresponding code repository\footnote{https://github.com/pix2pixzero/pix2pix-zero as of 04/03/2024.} mentions ``a large number of sentences ($\sim$1000)''.

Such a large number of captions is impractical for on-the-fly usage on a consumer-level GPU, since generating even a couple of sentences already takes a few seconds.
Taking this into account, we evaluate our approach along three practically viable scenarios, by asking the language model to generate one, two and four before-edit and after-edit captions.

\subsection{Zero-shot \vs few-shot}

LLMs seek to capture the patterns and structures that underlie the data they are trained on, which in many cases are different from the eventual task at hand.
Since LLMs often exhibit an inductive bias, meaning they have a tendency to reproduce certain patterns more readily than others, by providing some examples of how to perform a novel task one can bias the model towards reproducing the relevant patterns and features needed to succeed in that task.

Given this, we study the impact of giving none, one and three examples of how to perform the task to the model when asking it to generate the current before-edit and after-edit captions, that is zero-shot, 1-shot and 3-shot.


To create the pool of data from where to retrieve the examples for the few-shot prompts, we take 5\% of the development split of MAGICBRUSH and, for each such example, manually create four before-edit and four after-edit captions.
We randomly sample from this dataset to retrieve the few-shot examples given to the model in run time. When the model is asked to generate less than four captions, the few-shot examples should have the same number of captions, so we sample that amount of captions from the sampled example.

\subsection{Locking in the first before-edit caption}

Following \cref{sec:approach_step1}, Step~1 of our method requires a caption in order to perform DDIM Inversion on the initial image.
This caption is generated through the BLIP image captioning model.

Next, following \cref{sec:approach_step2}, in Step~2 the language model is prompted to generate a set of before-edit and after-edit captions.
The prompt used for this can be improved by giving, as the first before-edit caption, the caption already generated by BLIP in Step~1.
\cref{fig:prompt_lockin} gives an example, where ``A photo of an orange cat.'' is the caption generated by BLIP for the initial image.
Note that the edit request does not specify that the initial image is of a cat 
(\cf \cref{fig:prompt_template_filled}, where it does).



Locking in the first before-edit caption is especially useful when there is relevant information missing from the edit request.
For example, with the request ``make the green balloon a red balloon'' we know the entity that we begin with (``green balloon'') and the entity that we want to get (``red balloon'').
However, the request can be considerably more under-specified, such as ``make it a red balloon'', with no mention of the initial color of the balloon. Or it can be even more under-specified, as in ``make it red'', where all information about the entity is absent, and it is only know that we want it, whatever \emph{it} is, to be red.

Such under-specified requests highlight the usefulness of locking in the first before-edit caption.
Even if the request is something like ``make it red'', the caption provided by BLIP for the initial image (\eg~``a green balloon'') can provide relevant information to guide the generation of the captions that follow.

Considering this, we evaluate the impact of initializing the model with the first before-edit caption by re-using the caption obtained in Step~1. 
We further evaluate the impact that the quality of this caption has on the performance of the model, by utilizing the ``source'' gold caption provided in the MAGICBRUSH dataset instead of the caption generated by the BLIP model. Simulating a scenario where the caption is given by the user. 
This could have a strong impact especially if the captioning model fails to mention any of the entities in the initial image that are relevant to the edit request.

\section{Evaluation}
\label{sec:evaluation}


Following \cref{sec:data}, we evaluate our method on the MAGICBRUSH dataset~\cite{zhang:2024:magicbrush}, which comprises manually annotated image-editing triples.

In order to evaluate the performance of the various models in our experimental space, we adopt the cosine distance between CLIP embeddings.
We compare the embedding of the final edited image output by our model to both the embedding of the gold edited image (CLIP-I) and the embedding of the corresponding gold caption (CLIP-T) in the dataset.
We use CLIP ViT-B/32 to obtain these embeddings, which is the same model that was used in the MAGICBRUSH paper, as to ensure comparability with the scores reported there.

While comparing with the gold image seems an obvious choice, doing so with the gold caption might appear as a not-so-useful endeavour. 
Nevertheless, we realized that the latter option, that is comparing to the gold caption, is preferable as it permits a more fair performance scoring.

First, as mentioned in \cref{sec:data}, given the way MAGICBRUSH was created, sometimes there is information loss regarding the ``source'' image, such as the shape and color of entities, especially when these are not specified in the request (again, an example of this can be found in the contrast between the first and the fourth images in \cref{fig:dataset_example}).
While this is a problem when using the gold image in CLIP-I, this issue disappears when using the gold caption in CLIP-T because ``a black teddy bear'' is always a description of any black teddy bear despite its specific shape that happen to be represented in the gold image.

Second, DDIM Inversion introduces artifacts and removes or alters details present in the image.
By looking at the first and second images in \cref{fig:dataset_example}, it is possible to observe these differences.
Once again, this problem disappears when using CLIP-T, as details such as the positioning of the blades of grass or the fur pattern on the dog are not important to the distance scoring with it.

Finally, due to the nature of the task and following the example of \cref{fig:dataset_example}, when asked to ``make the teddy bear black'', there are various shades of black that could be used.
This is even more prevalent when the request is to add something that was not previously present in the image, since often the full details of what is to be added (its shape, color, position, \etc) are not specified.
When evaluated with the caption in CLIP-T, unless explicitly specified in the edit request, visual attributes such as shape, color, position and others do not influence the score.

\subsection{Comparing our models}
The performance results obtained with the various experiments are displayed in \cref{tab:results}, there the values in the ``avg''  columns are the average between scores obtain by the two studied prompts, and the ``stdev'' columns their standard deviation. In this table, it is possible to observe the following overall trends.
\begin{table}[tp]
\caption{CLIP cosine distance scores, averaged over two prompts, for each of the evaluated methods; best scores in bold.}
\label{tab:results}
\centering
\begin{tabular}{l @{\hspace{5ex}} cc @{\hspace{5ex}} cc}
\toprule
                           & \multicolumn{2}{l}{\hspace{3.6ex}CLIP-T}       & \multicolumn{2}{l}{\hspace{3.1ex}CLIP-I} \\
                           & avg              & stdev     & avg             & stdev \\
\midrule

\addlinespace
\multicolumn{5}{l}{Our model, 0-shot}                                                \\ 
\midrule
1-caption                  & 0.2751           & 0.0007    & 0.8021          & 0.0013 \\
1-caption; w. BLIP caption & 0.2795           & 0.0003    & 0.8255          & 0.0044 \\
2-caption; w. BLIP caption & 0.2796           & 0.0001    & 0.8329          & 0.0003 \\
4-caption; w. BLIP caption & 0.2799           & 0.0003    & 0.8347          & 0.0007 \\ 
\midrule

\addlinespace
\multicolumn{5}{l}{Our model, 1-shot}                                                \\ \midrule
1-caption                  & 0.2772           & 0.0012    & 0.8093          & 0.0023 \\
1-caption; w. BLIP caption & \textbf{0.2817}  & 0.0003    & 0.8310          & 0.0002 \\
2-caption; w. BLIP caption & 0.2800           & 0.0008    & 0.8328          & 0.0016 \\
4-caption; w. BLIP caption & 0.2797           & 0.0002    & 0.8348          & 0.0009 \\
\midrule

\addlinespace
\multicolumn{5}{l}{Our model, 3-shot}                                                \\
\midrule
1-caption                  & 0.2762           & 0.0003    & 0.8119          & 0.0032 \\
1-caption; w. BLIP caption & 0.2798           & 0.0001    & 0.8251          & 0.0001 \\
2-caption; w. BLIP caption & 0.2797           & 0.0000    & 0.8348          & 0.0010 \\
4-caption; w. BLIP caption & 0.2790           & 0.0000    & \textbf{0.8350} & 0.0011 \\
\bottomrule
\end{tabular}
\end{table}

First, when evaluating against the gold image in CLIP-I, the larger the number of examples in the few-shot prompt (from 0-shot to 3-shot) the better the performance; when evaluating against the gold text in CLIP-T, in turn, the same pattern is observed but only up to a certain point, as with 3-shot (containing three examples) the performance recedes with respect to the performance of 1-shot (containing only one example).

Second, locking in the first before-caption with the caption generated by BLIP in Step~1 permits better performance than having the language model generate it.
This provides significant strength to our claim in \cref{sec:experiments} that adding the first before-edit caption provides missing information to the model which is not present in the edit request.

Third, when evaluating against the gold image in CLIP-I, the larger the number of captions generated the better the performance. 
This improvement in scores can be attributed to a better definition of what represents the entity before and after the transformation, with more captions adding new information that better define these entities, as expected. 
When evaluating against the gold text in CLIP-T, in turn, the same pattern is observed but only in the 0-shot setting, as in the other few-shot settings the better performance is obtained with only one caption generated (provided it is the one generated by BLIP in Step~1).
As this is somewhat unexpected, we plan to research its cause in future work. 

Fourth, the intersection of these trends above leads to the best performance score of 0.2817 being the one obtained with 1-shot and the 1-caption generated by BLIP when taking the gold text in CLIP-T; while the best score of 0.8350 is obtained with 3-shot and 4-caption when taking the gold image in CLIP-I.

\subsection{Comparing with an oracle}

Following \cref{sec:evaluation}, and given the arguably superior fairness of the performance scores obtained with CLIP-T, we focus on the latter to evaluate the difference in performance between our models that are based on the caption generated by BLIP and those that are based on the gold source caption in MAGICBRUSH.
The results of this comparison is presented in \cref{tab:results_2}.

\begin{table}[tp]
\caption{CLIP cosine distance scores for the comparison between our best model and previous work. The $^*$ indicates values reported in the MAGICBRUSH paper~\cite{zhang:2024:magicbrush}.}
\label{tab:results_2}
\centering
\begin{tabular}{l @{\hspace{5ex}} cc @{\hspace{5ex}} cc}
\toprule
                          & \multicolumn{2}{l}{\hspace{3.5ex}CLIP-T}     & \multicolumn{2}{l}{\hspace{3ex}CLIP-I} \\
                          & avg          & stdev           & avg          & stdev       \\
\midrule
\multicolumn{5}{l}{Our best model, 1-shot, 1-caption (w/out preparation)}                                   \\ 
\midrule
\textit{with gold caption oracle}     & \textit{0.2845}   & \textit{0.0005}& \textit{0.8636}  & \textit{0.0015}  \\
w/out oracle              & \textbf{0.2817} & 0.0003& 0.8310         & 0.0002  \\ 
\midrule
\addlinespace
\multicolumn{5}{l}{Other models (w/ preparation)}                                                    \\ 
\midrule
InstructPix2pix$^*$            & 0.2764       &     -    & \textbf{0.8524}&    -    \\
HIVE$^*$                       & 0.2752       &     -    & 0.8519         &    -    \\
\midrule\midrule
\addlinespace
\multicolumn{5}{l}{Other models (w/ preparation also via fined-tuning on magicbrush)} \\ \midrule
 HIVE+magicbrush$^*$           & 0.2812       &     -    & 0.9189         &    -    \\ 
InstructPix2pix+magicbrush$^*$ & 0.2781       &     -    & 0.9332         &    -    \\
\bottomrule
\end{tabular}
\end{table}


As expected, our best model under CLIP-T evaluation, namely 1-shot and 1-caption with BLIP, performing at 0.2817, is surpassed by the model fully similar to it only that the BLIP caption is replaced by the gold caption of the source image in the MAGICBRUSH dataset, performing at 0.2845.

Interestingly, while this difference indicated the range of improvement that there still might be for our method, given this is a somewhat reduced difference, it indicates also that our method is already performing quite well.


\subsection{Comparing with previous work}

We turn now to the comparison with previous models reported in the literature. Unlike ours, these models underwent some preliminary preparation phase based on labelled datasets.
Among these, it is worth separating those that resorted also to the more sophisticated preparation by fine-tuning on MAGICBRUSH.

Considering first those that are not fine-tuned on the MAGICBRUSH data, and taking into account the scores in the arguably more fair distance to the CLIP-T test data, our best model emerges as outperforming all of them, with 0.2817 against the inferior 0.2764 (InstructPix2Pix) or 0.2752 (HIVE) scores.

Taking into account the dispreferred scoring distance that resorts to CLIP-I, in turn, when based in the oracle data, our best model emerges as outperforming all of them, with 0.8636 against the inferior 0.8524 (InstructPix2Pix) or 0.8519 (HIVE) scores.
This seems to indicate that even for the CLIP-I distance, there is room for our approach to be refined and eventually outperform the other models.

Finally, for reference purposes, the performance scores of models that have the advantage of being fine-tuned on the MAGICBRUSH data are also presented in the last two lines of Table \ref{tab:results_2}.
Interestingly, in spite of that clearly major advantage of supervised learning, our best model also outperform them, with 0.2817 against the inferior 0.2812 (HIVE+MAGICBRUSH) or 0.2781 (InstructPix2Pix+MAGICBRUSH) scores under the reliable CLIP-T distance.



\section{Discussion and Future Work}
\label{sec:discussion}

Taking into account the performance scores just presented, uur approach demonstrates to be highly competitive. when compared to existing methods in the literature, it outperforms all competitors in terms of the CLIP-T distance, falling short only when compared with the CLIP-I metric to supervised models fine-tuned specifically on the MAGICBRUSH dataset. The results reported demonstrate also that, despite this success, there remains room for improvement. 

Taking into account also the fact that our approach runs ``on the fly'' and that it was experimented on the basis of just out the shelf basic resources and settings, it turns out it not only demonstrated to be an highly effective method, but also a method with a high potential of progress.

First, the quality of the generated captions is a critical issue.
While Phi-2 performs admirably given its size, it lags behind larger models such as LLaMA2-65b~\cite{touvron:2023:llama2} or Falcon-180b~\cite{Almazrouei:2023:falcon}, not to mention models like GPT-3 and GPT-4~\cite{Brown:2020:GPT3,openai:2024:gpt4}. 

Second, as shown in our last experiment, enhancing the captioning quality will be crucial for improving the overall model performance and user satisfaction since it can mitigate the lack of information in edit requests.
Another interesting research avenue to alleviate this problem would be to resort to a chatbot-like system that would interact with the user to better understand the request.

In addition, the quality of image inversion was not thoroughly investigated in our paper, yet it can have a substantial impact on performance.
Employing techniques such as null prompt inversion \cite{mokady:2023:nullinversion} could prove beneficial in addressing this issue and improving model accuracy and robustness.

In future work, we aim to address these promising research paths and further enhance the capabilities of our model. This includes exploring advanced captioning techniques, leveraging larger language models, and investigating strategies to improve image inversion quality, among others. 

\subsection{Limitations and Potential Negative Impact}
\label{sec:limitations}

Image editing through natural language requests offers innovative opportunities for enhancing user engagement, creativity and accessibility. However, concerns arise regarding potential adverse effects due to the inherent nature of the technology.
Our methodology distinguishes itself by not requiring any form of training, yet it is important to acknowledge that we are subject to the limitations and potential biases of the pre-trained models we employ, namely Phi-2, Stable Diffusion, and BLIP. 
As in any technology, the manner in which it is used can significantly impact both its effectiveness and its ethical standing.

\section{Conclusion}
\label{sec:conclusion}

In this paper we presented a novel neural framework for image editing through natural language requests, which seamlessly integrates instruction-guided editing of images.
Our approach leverages three pre-trained models: Stable Diffusion, BLIP, and Phi-2, enabling users to modify images based on textual instructions without the need for preliminary specific training. 

We reported superior or highly competitive performance results for our proposal when compared to existing approaches in the literature, such as InstructPix2Pix \cite{brooks:2023:instructpix2pix} and HIVE \cite{zhang:2023:hive}, and obtained empirical evidence that it is an effective approach with a lot of potential for improvement in future research.

Additionally, the development of this technology can support a wide range of innovative applications, not only through its image editing capabilities but also by opening up new possibilities for individuals who may struggle with traditional interfaces. By enabling users to engage with images through natural language, it paves the way for enhanced accessibility and interaction.

%
%
\bibliographystyle{splncs04}
\bibliography{main}

\begin{thebibliography}{10}
\providecommand{\url}[1]{\texttt{#1}}
\providecommand{\urlprefix}{URL }
\providecommand{\doi}[1]{https://doi.org/#1}

\bibitem{Almazrouei:2023:falcon}
Almazrouei, E., Alobeidli, H., Alshamsi, A., et~al.: The {Falcon} series of language models: Towards open frontier models  (2023)

\bibitem{bahdanau:2014:attention}
Bahdanau, D., Cho, K., Bengio, Y.: Neural machine translation by jointly learning to align and translate. arXiv preprint arXiv:1409.0473  (2014)

\bibitem{brooks:2023:instructpix2pix}
Brooks, T., Holynski, A., Efros, A.A.: {InstructPix2Pix}: Learning to follow image editing instructions. In: Proceedings of the IEEE/CVF Conference on Computer Vision and Pattern Recognition. pp. 18392--18402 (2023)

\bibitem{Brown:2020:GPT3}
Brown, T.B., Mann, B., Ryder, N., et~al.: Language models are few-shot learners. arXiv preprint arXiv:2005.14165  (2020)

\bibitem{Cheng:2020:IIE}
Cheng, Y., Gan, Z., Li, Y., et~al.: Sequential attention {GAN} for interactive image editing. In: Proceedings of the 28th ACM International Conference on Multimedia. pp. 4383--4391 (2020)

\bibitem{Dhariwal:2021:diffusion}
Dhariwal, P., Nichol, A.: Diffusion models beat {GANs} on image synthesis. Advances in Neural Information Processing Systems  \textbf{34} (2021)

\bibitem{Dosovitskiy:2020:VIT}
Dosovitskiy, A., Beyer, L., Kolesnikov, A., et~al.: An image is worth 16x16 words: Transformers for image recognition at scale. arXiv preprint arXiv:2010.11929  (2020)

\bibitem{Galatolo:2021:CLIPGLASS}
Galatolo, F.A., Cimino, M.G., Vaglini, G.: Generating images from caption and vice versa via {CLIP}-guided generative latent space search. arXiv preprint arXiv:2102.01645  (2021)

\bibitem{Goodfellow:2014:GANs}
Goodfellow, I., Pouget-Abadie, J., Mirza, M., et~al.: Generative adversarial nets. Advances in Neural Information Processing Systems  \textbf{27} (2014)

\bibitem{gunasekar:2023:phi}
Gunasekar, S., Zhang, Y., Aneja, J., et~al.: Textbooks are all you need. arXiv preprint arXiv:2306.11644  (2023)

\bibitem{Guo:2018:shoesdataset}
Guo, X., Wu, H., Cheng, Y., et~al.: Dialog-based interactive image retrieval. In: Proceedings of the 32nd International Conference on Neural Information Processing Systems. pp. 676--686 (2018)

\bibitem{hertz:2022:prompt2prompt}
Hertz, A., Mokady, R., Tenenbaum, J., et~al.: Prompt-to-prompt image editing with cross attention control. arXiv preprint arXiv:2208.01626  (2022)

\bibitem{Hossain:2019:Captioning2}
Hossain, M.Z., Sohel, F., Shiratuddin, M.F., Laga, H.: A comprehensive survey of deep learning for image captioning. ACM Comput. Surv.  \textbf{51}(6) (feb 2019)

\bibitem{jiang:2023:mistral}
Jiang, A.Q., Sablayrolles, A., Mensch, A., et~al.: Mistral {7B} (2023)

\bibitem{Jiang:2021:LangGlobalEdit}
Jiang, W., Xu, N., Wang, J., et~al.: Language-guided global image editing via cross-modal cyclic mechanism. In: Proceedings of the IEEE/CVF International Conference on Computer Vision. pp. 2115--2124 (2021)

\bibitem{Jiang:2021:TransGAN}
Jiang, Y., Chang, S., Wang, Z.: {TransGAN}: Two transformers can make one strong {GAN}  (2021)

\bibitem{Kovashka:2012:relative2}
Kovashka, A., Parikh, D., Grauman, K.: Whittlesearch: Image search with relative attribute feedback. In: 2012 IEEE Conference on Computer Vision and Pattern Recognition. pp. 2973--2980. IEEE (2012)

\bibitem{Lecun:1989:CNN}
LeCun, Y., Boser, B., Denker, J., et~al.: Handwritten digit recognition with a back-propagation network. Advances in Neural Information Processing Systems  \textbf{2} (1989)

\bibitem{li:2022:blip}
Li, J., Li, D., Xiong, C., Hoi, S.: Blip: Bootstrapping language-image pre-training for unified vision-language understanding and generation. In: International Conference on Machine Learning. pp. 12888--12900. PMLR (2022)

\bibitem{lin:2014:coco}
Lin, T.Y., Maire, M., Belongie, S., et~al.: Microsoft {COCO}: Common objects in context. In: Proceedings of ECCV 2014: 13th European Conference on Computer Vision. pp. 740--755. Springer (2014)

\bibitem{mokady:2023:nullinversion}
Mokady, R., Hertz, A., Aberman, K., et~al.: Null-text inversion for editing real images using guided diffusion models. In: Proceedings of the IEEE/CVF Conference on Computer Vision and Pattern Recognition. pp. 6038--6047 (2023)

\bibitem{openai:2024:gpt4}
OpenAI: {GPT-4} technical report (2024)

\bibitem{parmar:2023:pix2pix-zero}
Parmar, G., Kumar~Singh, K., Zhang, R., et~al.: Zero-shot image-to-image translation. In: ACM SIGGRAPH 2023 Conference Proceedings. pp. 1--11 (2023)

\bibitem{Radford:2021:CLIP}
Radford, A., Kim, J.W., Hallacy, C., et~al.: Learning transferable visual models from natural language supervision. arXiv preprint arXiv:2103.00020  (2021)

\bibitem{Ramesh:2022:DALLE2}
Ramesh, A., Dhariwal, P., Nichol, A., et~al.: Hierarchical text-conditional image generation with {CLIP} latents. arXiv preprint arXiv:2204.06125  (2022)

\bibitem{Ramesh:2021:DALLE}
Ramesh, A., Pavlov, M., Goh, G., et~al.: Zero-shot text-to-image generation. arXiv preprint arXiv:2102.12092  (2021)

\bibitem{Reed:2016:Generative}
Reed, S., Akata, Z., Yan, X., et~al.: Generative adversarial text to image synthesis. In: International Conference on Machine Learning. pp. 1060--1069. PMLR (2016)

\bibitem{rombach:2022:stable_diffusion}
Rombach, R., Blattmann, A., Lorenz, D., et~al.: High-resolution image synthesis with latent diffusion models. In: Proceedings of the IEEE/CVF Conference on Computer Vision and Pattern Recognition. pp. 10684--10695 (2022)

\bibitem{song:2020:DDIM}
Song, J., Meng, C., Ermon, S.: Denoising diffusion implicit models. arXiv preprint arXiv:2010.02502  (2020)

\bibitem{sutskever2014sequence}
Sutskever, I., Vinyals, O., Le, Q.V.: Sequence to sequence learning with neural networks. Advances in Neural Information Processing Systems  \textbf{27} (2014)

\bibitem{tao2021dfgan}
Tao, M., Tang, H., Wu, S., et~al.: {DF-GAN}: Deep fusion generative adversarial networks for text-to-image synthesis (2021)

\bibitem{touvron:2023:llama2}
Touvron, H., Martin, L., Stone, K., et~al.: {LLaMA~2}: Open foundation and fine-tuned chat models. arXiv preprint arXiv:2307.09288  (2023)

\bibitem{Vaswani:2017:Transformer}
Vaswani, A., Shazeer, N., Parmar, N., et~al.: Attention is all you need. In: Advances in Neural Information Processing Systems. pp. 5998--6008 (2017)

\bibitem{wu:2021:NUWA}
Wu, C., Liang, J., Ji, L., et~al.: {N\"UWA}: Visual synthesis pre-training for neural visual world creation. arXiv preprint arXiv:2111.12417  (2021)

\bibitem{Wu:2017:Captioning1}
Wu, Q., Shen, C., Wang, P., et~al.: Image captioning and visual question answering based on attributes and external knowledge. IEEE Transactions on Pattern Analysis and Machine Intelligence  \textbf{40}(6),  1367--1381 (2017)

\bibitem{Xu:2015:Captioning}
Xu, K., Ba, J., Kiros, R., et~al.: Show, attend and tell: Neural image caption generation with visual attention. In: International Conference on Machine Learning. pp. 2048--2057. PMLR (2015)

\bibitem{xu2017attngan}
Xu, T., Zhang, P., Huang, Q., et~al.: {AttnGAN}: Fine-grained text to image generation with attentional generative adversarial networks (11 2017)

\bibitem{yang:2023:llm_optimizers}
Yang, C., Wang, X., Lu, Y., et~al.: Large language models as optimizers. arXiv preprint arXiv:2309.03409  (2023)

\bibitem{Yu:2017:relative1}
Yu, A., Grauman, K.: Fine-grained comparisons with attributes. In: Visual Attributes, pp. 119--154. Springer (2017)

\bibitem{zhang:2024:magicbrush}
Zhang, K., Mo, L., Chen, W., et~al.: Magicbrush: A manually annotated dataset for instruction-guided image editing. Advances in Neural Information Processing Systems  \textbf{36} (2024)

\bibitem{zhang:2023:hive}
Zhang, S., Yang, X., Feng, Y., et~al.: {HIVE}: Harnessing human feedback for instructional visual editing. arXiv preprint arXiv:2303.09618  (2023)

\bibitem{Zhu2019DMGANDM}
Zhu, M., Pan, P., Chen, W., Yang, Y.: {DM-GAN}: Dynamic memory generative adversarial networks for text-to-image synthesis. 2019 IEEE/CVF Conference on Computer Vision and Pattern Recognition (CVPR) pp. 5795--5803 (2019)

\bibitem{Zhuang:2021:EditingLatentSpace}
Zhuang, P., Koyejo, O., Schwing, A.G.: Enjoy your editing: Controllable {GANs} for image editing via latent space navigation. arXiv preprint arXiv:2102.01187  (2021)

\end{thebibliography}
\end{document}